\documentclass[11pt]{article}

\usepackage[final]{acl}

\usepackage{times}
\usepackage{latexsym}
\usepackage[T2A,T1]{fontenc}
\usepackage[utf8]{inputenc}
\usepackage{microtype}
\usepackage{inconsolata}
\usepackage{graphicx}
\usepackage{amsmath}
\usepackage{booktabs}
\usepackage{lineno}
\usepackage{subcaption}
\usepackage{float}
\usepackage{tabularx}
\usepackage{array}

\title{Entropy of Ukrainian}

\author{Anton Lavreniuk$^{1}$ \and Mykyta Mudryi$^{1,2}$ \and Markiian Chaklosh$^{1,3}$ \\
$^{1}$ARIMLABS.AI \quad $^{2}$Polish-Japanese Academy of Information Technology \\
$^{3}$University of the National Education Commission in Kraków \\
\texttt{\{alavreniuk, mmudryi, mchaklosh\}@arimlabs.ai}}

\begin{document}
\maketitle
\begin{abstract}
        In natural language processing, the entropy of a language is a measure of its unpredictability and complexity. The first study on this subject was conducted by Claude Shannon in 1951. By having participants predict the next character in a sentence, he was able to approximate the entropy of the English language. Several follow-up studies by other authors have since been conducted for English, and one for Hebrew. However, to date, Shannon's experiment has never been conducted for Ukrainian. In this paper, we perform this experiment for Ukrainian by recruiting 184 volunteers using social media channels. We rely on techniques used for English to approximate the entropy value of Ukrainian. The final result is an upper bound of $H_{upper}\approx1.201$  bits per character. We compare this to the performance of current Large Language Models. The methods and code used are also documented and published, along with a discussion of the main challenges encountered. 
\end{abstract}

\section{Introduction}
In information theory, entropy is a measure of the information content or surprise associated with an event. 

For a natural language, entropy represents the unpredictability of its characters. Entropy is a fundamental property of the language itself.
Measuring this value has broad implications, including but not limited to cross-linguistic information density comparisons, data compression, and language modeling. 

This unpredictability cannot be calculated directly, as that would require knowing the true probability distribution of the language - which words follow which, which topics relate to each other, all for arbitrarily long contexts and across all speakers of that language. 

Several methods for approximating the entropy of a language have been used. 

Statistical N-gram-based models produce exact values for their corpora but are greatly limited by data sparsity as N grows. One such study was conducted for Ukrainian \citep{babenko2012entropy}, estimating the entropy value for N values up to 9, with the result $H_9 = 1.25\text{--}1.40$ bpc. 

Another method is compression-based analysis. \citet{takahira2016entropy} attempted to use PPM compression on massive corpora for entropy estimation, with the smallest bound for English being $H \approx 1.4$ bpc. However, this method consistently overestimates entropy due to the limited predictive capacity and restrictive context windows of compression algorithms. 

Some neural-network approaches have been used to attempt to measure language complexity. They have the advantage of being able to access the model's internal probabilities and calculate entropy directly, avoiding bounds-based calculations. \citet{takahashi2018cross} measured the complexity of English with neural networks, resulting in $H \approx 1.12$ bpc. However, these methods rely on minimizing the cross-entropy loss for a training dataset 

\begin{linenomath*}
\[D_{training} \subset D_{language}\]
\end{linenomath*}

, which means that the resulting entropy estimate reflects the chosen training corpus instead of the entire language. This makes overfitting to a small or unrepresentative corpus a significant concern. Humans are the source of ground truth for the natural language they speak, and as such are much less prone to overfitting. Modern Transformer-based language models perform significantly better than humans, but their training corpora might not represent their language fully, leaving gaps or overfocusing on some aspects of a language. 

A different approach, using human knowledge, was proposed in 1951 by Claude Shannon \citep{shannon1951prediction}.

By making humans guess the next character in a sentence, it is possible to closely approximate a lower and upper bound on the true entropy of a natural language using the number of guesses it takes a human to guess the next character given N characters of context. Note that calculating the exact number would require knowing the exact probabilities that a human assigns to guesses in their head, which is impossible. One study \citep{cover1978convergent} asked humans to output the entire probability distribution of their guesses. However, while theoretically improving the results, this severely strains human participants who are forced to set a percentage probability for each of the 27 characters. Additionally, humans' internal probabilities are likely different from what a human can actually output, as humans are not consciously aware of the exact computations performed by their brains.

As such, later studies used the method originally described in \citet{shannon1951prediction}, and aimed at increasing the number of participants to combat bias.

After considering the different methods available, we chose Shannon's experiment for estimating the entropy of Ukrainian - the first time this experiment has been performed for this language.

A section is also dedicated to using LLMs for entropy estimations.

\section{Methodology}

We conducted Shannon's experiment, based on the methodology described in \citet{ren2019entropy}. Due to the volunteer-based nature of the experiment, extensive modifications were made to the experimental setting to ensure participant engagement and minimize frustration and early quitting. Namely, the volunteers were provided with 70 characters of initial context as opposed to starting from 0, and the dataset was filtered to sentences from 120 to 200 characters long. The user interface included a small delay after each attempt to prevent button mashing. Users could complete as many sessions as they wanted.

\subsection{Participant Recruitment}

Volunteers were primarily recruited through the Telegram messenger and word-of-mouth, with the pool consisting mostly of young adults and adults willing to take part in the experiment. The recruitment was conducted in Ukrainian, with participants expected to be native Ukrainian speakers, though language background was not formally assessed. A giveaway was held to encourage participation, with victory odds being directly proportional to the number of sessions completed, and at least 2 completed sessions required to be eligible. The total experiment cost was $\sim$200 USD, split between giveaway prizes and website hosting. A total of 323 people started the registration process and 256 confirmed their intention to participate. Out of those, 184 started at least one session and 131 completed at least two sessions. The total number of completed sessions was 501, and data from 192 more incomplete sessions were also used in the final calculations. Overall participant engagement was moderate, with the median volunteer completing 2 sessions. 
 
\begin{figure}[t]
\centering
\includegraphics[width=0.7\columnwidth]{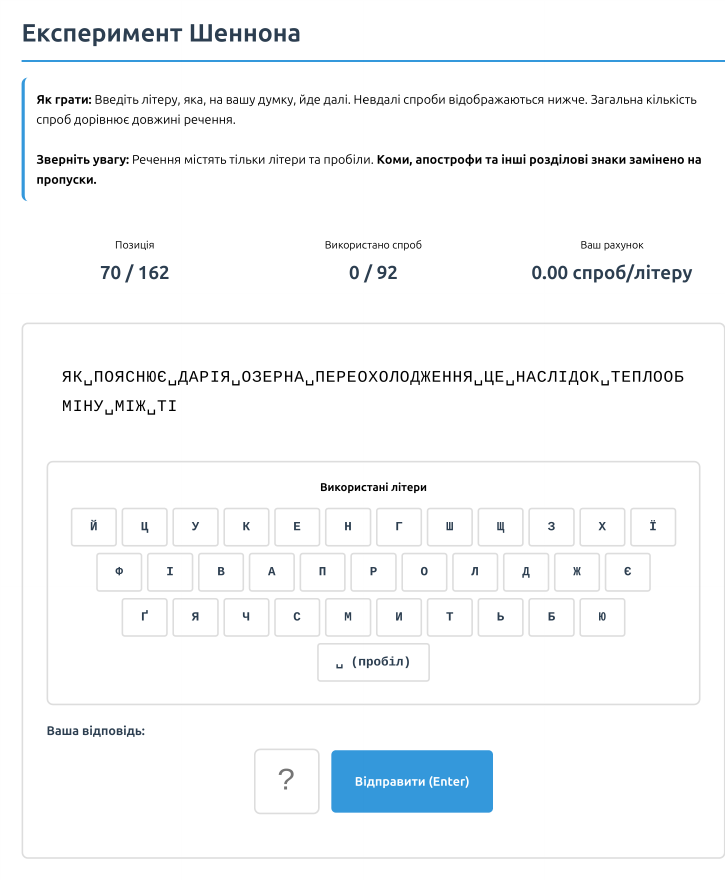}
\caption{User interface used during the experiment}
\label{fig:ui}
\end{figure}

Following the methodology of \citet{ren2019entropy}, who used news articles, our dataset consisted of 136 sentences collected from 5 news articles on different topics from Ukrainska Pravda, published from August 8, 2025 to January 23, 2026. The pre-processing of articles consisted of splitting into sentences on ".!?" and filtering out sentences that contained Latin characters or digits. All characters that are not whitespace or one of the 33 letters of the Ukrainian alphabet were replaced with whitespace. Lastly, all neighboring whitespaces were replaced with a single whitespace, and sentences shorter than 120 or longer than 200 characters were discarded. 

For each session, a volunteer is presented with one sentence from the total sentence pool, with the initial 70 characters revealed, and prompted to guess the next character. They keep guessing that character until a correct one is entered, which constitutes a single observation - the number of attempts needed to guess a character at position N. For example, a participant seeing the sentence {\fontencoding{T2A}\selectfont "ЯК\_ ПОЯСНЮЄ\_ ДАРІЯ\_ ОЗЕРНА\_ ПЕРЕОХОЛОДЖЕННЯ\_ ЦЕ\_ НАСЛІДОК\_ ТЕПЛООБМІНУ\_ МІЖ\_ ТІ"}(*YAK POIASNIUIE DARIIA OZERNA PEREOKHYLODZHENNIA TSE NASLIDOK TEPLOOBMINU MIZH TI* `as Dariia Ozerna explains, hypothermia is a consequence of heat exchange between [the] bo-') might guess the character {\fontencoding{T2A}\selectfont "Н"}(*N*) incorrectly, having it highlighted in red, then guess the character {\fontencoding{T2A}\selectfont "Л"}(*L*)correctly, resulting in gathering an observation of "2 guess attempts until correct at 70 characters of context", and the participant proceeding to guessing the next character.  

The length of a session is limited by the number of guesses available. For each sentence, the number of total guess attempts available is equal to its total length minus the 70 revealed characters. As such, completing the sentence would require guessing each character correctly on the first try, and this was not expected to occur in practice. A session ends when all the attempts are exhausted.

After gathering a sufficient number of observations, the entropy bounds are calculated as follows.

Given $K$ as the number of possible guesses (alphabet size of the target language + whitespace), and $q_i$ as the fraction of characters guessed correctly on the $i$-th try, the true entropy $H$ is approximated by 
\begin{linenomath*}
\[H_{lower} \leq H \leq H_{upper}\]
\end{linenomath*}

, where the upper bound is the entropy of the distribution of guess counts 

\begin{linenomath*}
\[H_{upper} = -\sum_{i=1}^{K} q_i \log_2 q_i\]
\end{linenomath*}

    and the lower bound is the entropy of the "least surprising" distribution - one where each character guessed on the $i$-th try had a probability of $\frac{1}{i}$ 
    
\begin{linenomath*}
\[H_{lower} = \sum_{i=1}^{K} q_i \log_2 i\]
\end{linenomath*}

    \citep{shannon1951prediction}
    
\section{Result Analysis}

In this analysis, we follow methods used in \citet{ren2019entropy}.  

Data collection took place from January 24 to January 31, 2026. During this period, 853 total sessions were started, out of which 501 were fully completed (ran out of guesses) and 352 were abandoned early. Note that guesses from abandoned sessions are also included in the final calculations. The completed sessions contributed 38,977 character guesses and the unfinished sessions contributed 5788 guesses for a total of 44,765 character guesses. Out of those, 17,023 guesses (38\%) were correct. These 17,023 correct guesses correspond to 17,023 observations - $\sim$10\% of Ren et al.'s scale of 172,954 observations. The number of observations is highest for positions 70-90, and drops off sharply after 100 as participants exhaust their guesses. We discard positions after 110 due to insufficient data.

\begin{figure}[t]
    \centering
    \includegraphics[width=0.7\columnwidth]{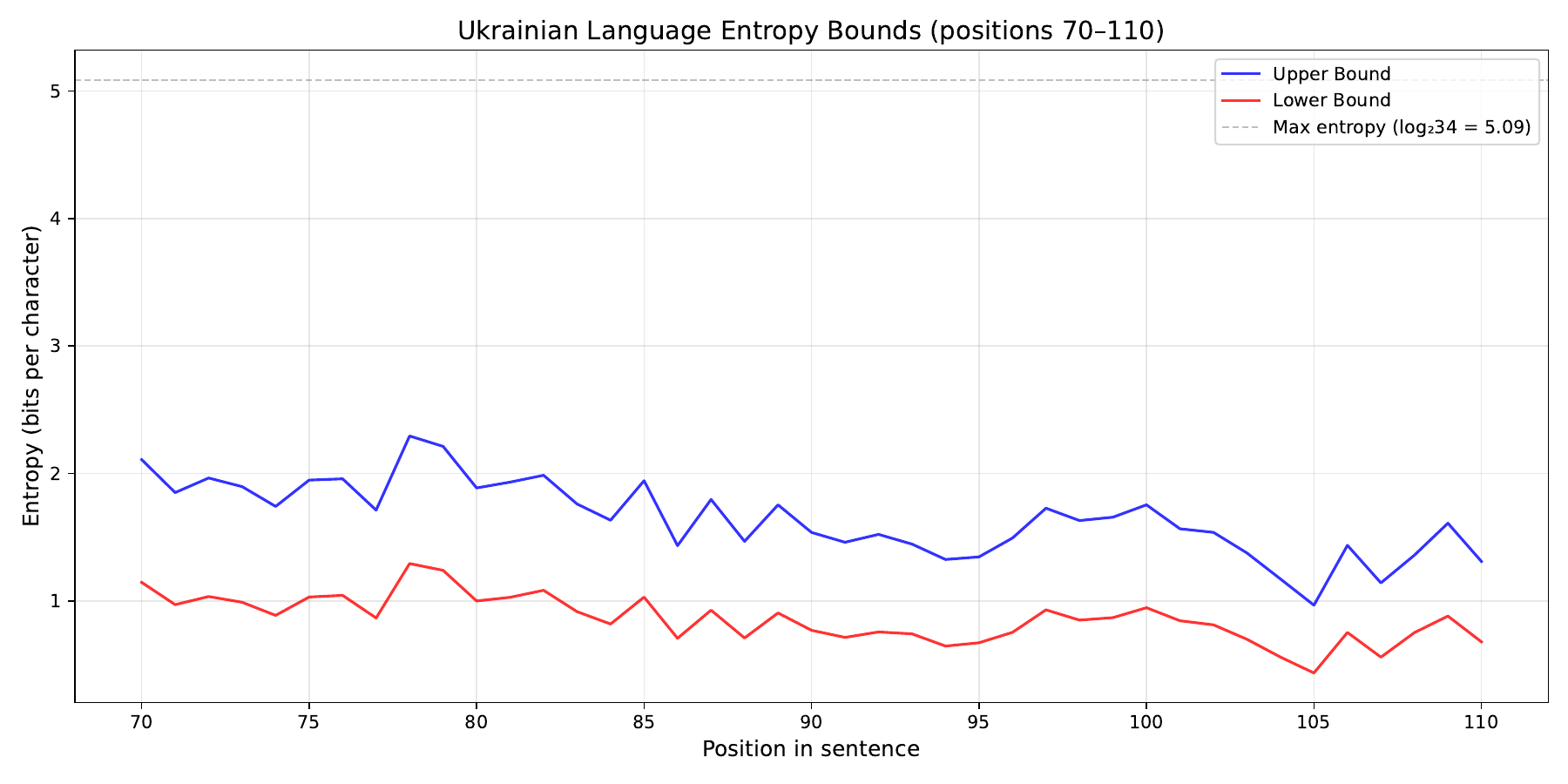}
    \caption{Lower and Upper bounds of entropy per position for positions 70-110}
    \label{fig:entropybypos}
\end{figure}

\begin{figure}[t]
    \centering
    \includegraphics[width=0.7\columnwidth]{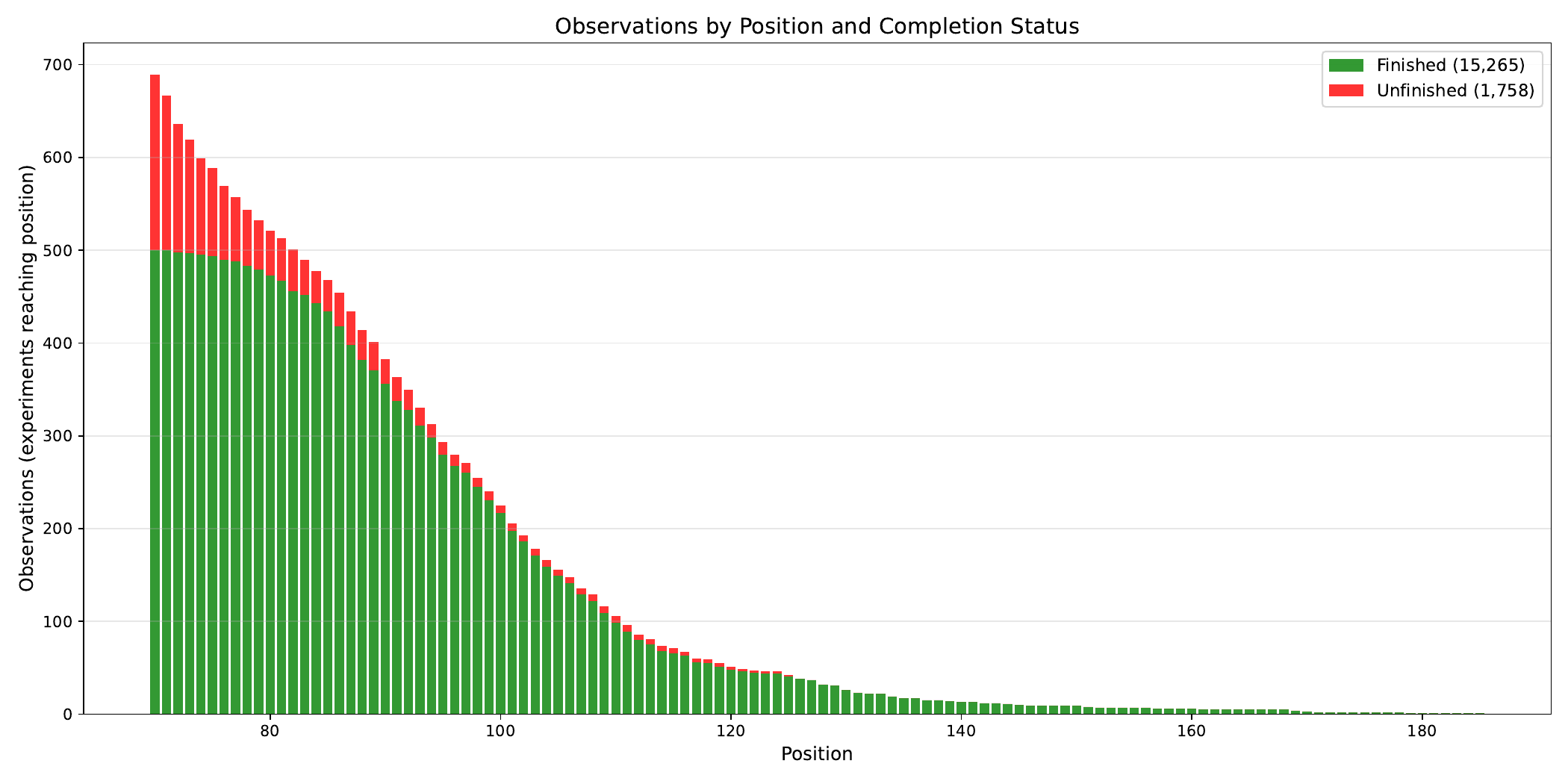}
    \caption{Number of observations gathered per position}
    \label{fig:obsbypos}
\end{figure}

The per-position entropy values exhibit considerable variance. A pooling approach, together with filtering, is used to mitigate this.

\subsection{Constraining Positions}

\citet{ren2019entropy} relied on the application of an ansatz function to fit the existing data and extrapolate the entropy value to infinite context length. This approach requires data starting from position 0, which was not gathered due to participant engagement concerns. Alternative curve-fitting approaches were attempted but were unreliable due to noise in the collected data.
Instead, we use the fact that context beyond $N=70$ only marginally improves guess accuracy. The value of the ansatz function from \citet{ren2019entropy} at $N=70$ differs from the extrapolated value by approximately 1\%. While this ansatz value may not perfectly model English or Ukrainian, calculating an estimate using only positions 70-110, without extrapolation, is highly unlikely to introduce more error than other sources of variance in this experiment. Additionally, guess attempt limits bias the data by filtering out poor performers early and populating the larger N values only by observations from skilled guessers.

\subsection{Data Trimming}

For the purposes of calculating an upper bound on natural language entropy, we are interested in using only the results from top performers, as their results are an indication of low natural language entropy, and poor performance by a participant might indicate a lack of commitment or interest as opposed to language complexity. 
As such, we need to trim the existing data.

\citet{ren2019entropy} trimmed the 50\% of the worst-performing sessions, dropping the final estimate by about 0.2 bpc. The goal is to discard poor performers without removing the natural variance in responses. However, this makes the final estimate more sensitive to cheating by top performers in online experimental settings. 

We perform additional outlier detection beforehand to discard improbably good results. By running a binomial analysis and discarding sessions which had a $<1\%$ chance of arising given the mean accuracy score, we discard 30 sessions $(4.3\%)$, marked as suspicious.
This accounts for possible random guessing, as well as possible noise or cheating, as some participants reported that they were able to look up the full sentences online.

Additionally, we argue that volunteer participants are not as engaged as paid MTurk workers. We revise the trimming percentage to remove 65\% of the worst performing humans. This trim is chosen due to having one of the smallest CI widths, and due to the insufficiency of the original 50\% trim for volunteer participants.

Note that the choice of trim matters greatly for the final result. In Table~\ref{tab:trim_sensitivity}, the effect of trim percentage on the final result is shown. Choosing a trim percentage is a balance of excluding underperforming participants and not removing true performers in favor of lucky guesses. More precise results can be achieved by using controlled environments and unpublished datasets (removing cheating), and additional participant encouragement, such as direct monetary compensation, participant filtering, and directly rewarding better results (removing underperformance), as well as larger sample sizes.

\begin{table*}                                                                                                                                                                                                                                                   
\centering                                                                                                                                                                                                                                                           
\caption{Sensitivity of the upper bound to bottom-trim level after binomial outlier removal. Session-level bootstrap, 2{,}000 iterations.}                                                                                                                              
\label{tab:trim_sensitivity}
\begin{tabular}{rrcccc}                                                                                            
\toprule                                                                                                
Bottom trim & Pool & Point est. (bpc) & Bootstrap median & 95\% CI & Width \\
\midrule
0\%  & 663 & 1.830 & 1.790 & [1.731, 1.849] & 0.118 \\
10\% & 597 & 1.756 & 1.714 & [1.656, 1.764] & 0.109 \\
20\% & 531 & 1.671 & 1.630 & [1.582, 1.682] & 0.099 \\
30\% & 465 & 1.581 & 1.536 & [1.485, 1.584] & 0.099 \\
40\% & 398 & 1.493 & 1.443 & [1.389, 1.492] & 0.103 \\
50\% & 332 & 1.382 & 1.331 & [1.279, 1.373] & 0.094 \\
55\% & 299 & 1.327 & 1.280 & [1.233, 1.322] & 0.088 \\
60\% & 266 & 1.255 & 1.204 & [1.154, 1.246] & 0.092 \\
\textbf{65\%} & \textbf{233} & \textbf{1.201} & \textbf{1.149} & \textbf{[1.102, 1.192]} & \textbf{0.090} \\
70\% & 199 & 1.136 & 1.083 & [1.029, 1.128] & 0.098 \\
80\% & 133 & 0.896 & 0.839 & [0.753, 0.887] & 0.134 \\
90\% &  67 & 0.327 & 0.290 & [0.211, 0.352] & 0.141 \\
\bottomrule
\end{tabular}
\end{table*}

\subsection{Entropy Calculation}

For a more robust entropy calculation, observations from positions 70-110 are pooled. This brings the overall number of observations used for this calculation post-trimming to 4,869.

The final upper entropy bound of the collected observations is derived from a mean of all upper entropy bounds in range 70-110, weighted by the number of observations for that position. 

The formula used is 
\begin{linenomath*}
\[H_{upper} = \sum_{n=n_1}^{n_2} w_n \cdot H_{upper}(n), \quad w_n = \frac{N_n}{\sum_{k} N_k}\]
\end{linenomath*}

, where $n_1 = 70, n_2=110$, and $N_n$ is the number of observations for position $n$, and $H_{upper}(n)$ is the upper bound calculated for position $n$.

The resulting upper bound is $H_{upper} = 1.201$ bpc. 

The lower entropy bound, calculated in the same way for $H_{lower}(n)$, is $H_{lower} = 0.5987$ bpc.

\begin{figure}[t]
    \centering
    \includegraphics[width=0.7\columnwidth]{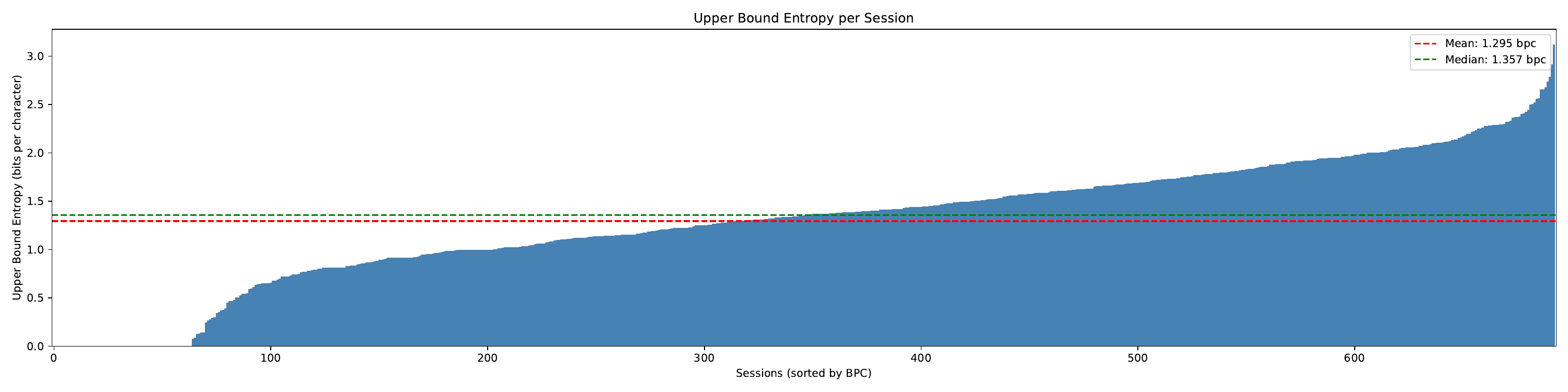}
    \caption{BPC per session for all sessions, ordered by performance.}
    \label{fig:bpcpersession}
\end{figure}

\subsection{Bootstrap Analysis}
To quantify the uncertainty in this result, we perform a bootstrap analysis on the existing data. We resample sessions with replacement 2000 times to construct a bootstrap distribution of the upper bound estimate. The 95\% CI is  [1.10, 1.19], with a width of 0.090. The downward bias of the bootstrap median compared to the initial point estimate is consistent with small sample sizes \citep{ren2019entropy}. Note that the final calculations used 233 different sessions post-trim, which is approximately 23\% of the 1000 sessions per N used for calculations in \cite{ren2019entropy}. Increasing sample size is expected to reduce confidence intervals further. Direct comparisons to \cite{ren2019entropy} are not possible due to only pre-trimming 90\% CIs being published by the authors.

\subsection{Final Result}

We report an upper bound on entropy for Ukrainian of 1.201 bpc, a redundancy of $76.4\%$. 

Per \citet{babenko2012entropy}, conditional n-gram entropy of Ukrainian is $1.25-1.40$ bpc for $N=9$. This represents a more direct computation limited to short context windows. Conditional calculations for longer N values are impossible due to the frequency of n-grams diminishing rapidly. Our result using human participants confirms that the trend of lesser entropy with larger context windows continues beyond N=9 for Ukrainian.

\subsection{Comparison to English}

Comparing to the results of \citet{ren2019entropy} for English, we see very similar patterns. 

Their experiment consisted of predicting 27 characters, 26 from the English alphabet and whitespace, with the max entropy of $\log_2(27)\approx4.75bpc$. For Ukrainian, we use 34 characters, 33 letters of the alphabet + whitespace, with the max entropy of $\log_2(34)\approx5.09bpc$.

Despite this, the entropy and redundancy values of $\approx1.20bpc$ + 76.4\% for Ukrainian and $\approx1.22bpc$ + 74.3\% for English are similar. It is likely that a more refined measurement for either language will result in lower estimates. Despite Ukrainian having 7 more letters, some are seldom used, contributing less to increasing the overall entropy value. 

Both experiments suffered from an online setting, having some shared biases in the form of cheating and inconsistent participant engagement. Additionally, our trimming was more aggressive due to the use of volunteer participants.

In general, there are no substantial differences in the entropy estimates between the two languages. 

\section{Large Language Models as Predictors of Ukrainian}

We compare human results to the performance of recent Transformer-based Large Language Models. 

As outlined in Section 1, neural-network based methods provide lower entropy estimates than other approaches. The core concern that disallows using those estimates as true language entropy bounds is overfitting the test domain. Nevertheless, we provide results from LLM entropy measurements on the same data corpus, effectively comparing LLM performance to human results. 

Given a language model $M$. For a string of N characters long text, a token-level language model tokenizes it into T tokens. Given T, BPC is equal to: 

\begin{linenomath*}
\[\text{BPC} = -\frac{1}{N} \sum_{i=1}^{T} \log_2 P_M(t_i \mid t_1, t_2, \ldots, t_{i-1})\]
\end{linenomath*}

, where $t_i$ is the i-th token in the sequence, and $P_M(t_i \mid t_1, t_2, \ldots, t_{i-1})$ is the probability assigned to it by the model given i-1 preceding tokens. Dividing by N normalizes the value and removes tokenizer differences, allowing for direct comparisons.

In practice, language models add an additional cost from imperfect language modeling. For a language model, the BPC is equal to $\text{BPC} = H(p) + D_{\text{KL}}(p \| q) \geq H(p)$, where p is the true probability distribution of a language, q is the learned distribution, $H(p)$ is the entropy of the language, and $D_{\text{KL}}(p \| q)$ is the Kullback–Leibler divergence between the two distributions, which is the result of imperfect modeling. KL divergence is the number that is reducible by better training, and is the main metric that we are interested in when comparing models. 

For measuring language knowledge, we use pre-trained model checkpoints before instruction tuning takes place. Fewer pre-processing steps are performed - characters are not capitalized and punctuation is not removed. We only count loss on tokens that start after position 70. 

To prevent data contamination, the news dataset spans August 2025 - January 2026, after the training cutoffs for LLMs being benchmarked.
\begin{figure}
    \centering
    \includegraphics[width=1\linewidth]{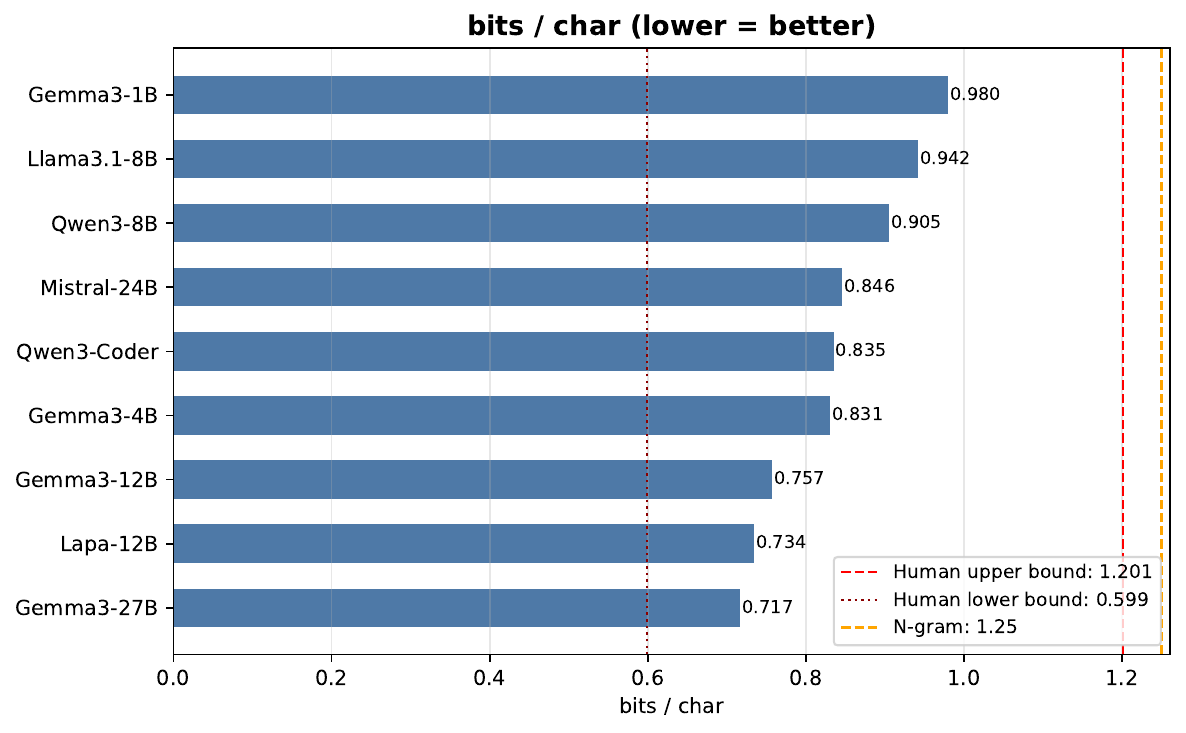}
    \caption{LLM Results for Shannon's experiment}
    \label{fig:llmbpc}
\end{figure}

\begin{table*}[t]
\centering
\caption{LLM results on the experiment corpus, sorted by BPC.}
\label{tab:llm_results}
\begin{tabular}{lrrcc}
\toprule
Model & Params & Fertility (Chars/Token) & BPC \\
\midrule
Gemma-3          & 27B  & 3.33 & 0.717 \\
Lapa             & 12B  & 5.67 & 0.734 \\
Gemma-3          & 12B  & 3.33 & 0.757 \\
Gemma-3          & 4B   & 3.33 & 0.831 \\
Qwen3-Coder-Next & 80B  & 2.03 & 0.835 \\
Mistral-Small    & 24B  & 2.95 & 0.846 \\
Qwen3            & 8B   & 2.03 & 0.905 \\
Llama-3.1        & 8B   & 3.24 & 0.942 \\
Gemma-3          & 1B   & 3.33 & 0.980 \\
\midrule
Human upper bound & --  & --   & 1.201 \\
Human lower bound & --  & --   & 0.599 \\
\bottomrule
\end{tabular}
\end{table*}

LLM results show substantial improvements over other measurement types. Llama 3.1, which does not officially support Ukrainian, scores worse than other models which do support Ukrainian. Gemma 27B and Lapa~\citep{lapa2025}, a Gemma 12B finetune on Ukrainian, show the best results, suggesting low KL divergence. Lapa has the highest fertility due to its updated tokenizer, however, its good results are not directly attributable to this. Many models get close to the human lower bound, suggesting that overfitting to the news domain is present. Note that those results were measured with sentence-level context windows of 120-200 characters, mirroring the main experiment. Gemma-3 scales cleanly across parameter counts (1B→4B→12B→27B), with BPC improving from 0.980 to 0.717

The relationship of fertility to BPC is not conclusive - Qwen models show decent results on Ukrainian despite fertility of $\approx2$ characters per token. Establishing a relationship of fertility to bpc requires follow-up investigation.
All the obtained results are well below \citep{takahashi2018cross}'s 1.12 BPC result, due to using much newer and larger transformer-based models.

The best results being in the $\approx 0.7$BPC range imply that the real language entropy lies below our human experiment results. However, results being very close to the human lower bound suggest that overfitting is present. Quantifying the degree of overfitting is not trivial and requires follow-up investigation, however, modern language models might be a pathway to measuring entropy of natural language.

\section{Conclusion}

This paper presented the results of a replication of Shannon's experiment for the Ukrainian language. Relying on methods from \citet{ren2019entropy}, we recruited volunteers and gathered 17,023 observations from 184 individuals. This was the first time this experiment was performed for the Ukrainian language. After extensive analysis and data filtering due to the online environment of the experiment, we calculate the upper bound of Ukrainian given at least 70 but no more than 110 characters of context to be $\approx 1.20 bpc$, and, based on the ansatz function from \citet{ren2019entropy}, argue that this value is close to the entropy bound given infinite context. This improves upon the only previous estimate for Ukraine by \citet{babenko2012entropy}, whose n-gram-based approach yielded $>1.25 bpc$. The final results are compared to English, showing similar information densities, and to LLM performance, showing a greater predictive performance in language models. We describe the limitations of this approach and the difficulties in using volunteer participants, and point out directions for future work. We also publish code and data used in the experiment.

\section*{Limitations}

The primary limitation of this experiment is the volume of data gathered. 

184 participants contributed a total of 17,023 total and 4,869 post-filtering observations, which proved to be insufficient for a precise result. Comparing to \citet{ren2019entropy}, a volunteer on average gathered 93 observations, compared to 253 observations for a paid MTurk worker. A volunteer is naturally less engaged than a paid worker, leading to lower accuracy and, consequently, higher estimates. Additionally, our experiment started from position 70 to ensure participant engagement. This makes some analysis, such as fitting an ansatz function or plotting the increase in precision with more context, impossible.

Additionally, the choice of a dataset influences the results greatly. Some parts of a language, such as informal speech, regional variations, and personal accounts of unknown events will naturally have much higher unpredictability compared to widely-known idioms, proverbs and works of art, which can be not only predicted but recited by most native speakers. We follow \citet{shannon1951prediction} and \citet{ren2019entropy}, which used datasets composed of news articles, with 100 and 225 sentences respectively. Our dataset consisted of 136 sentences sampled from 5 recent articles, which likely introduced bias due to limited topical and stylistic variety.

\subsection*{Future Work}

For future work, we recommend recruiting at least 500 participants, and gathering at least 100,000 observations, with context lengths ranging from zero to several hundred characters, allowing for extrapolation via fitting of an ansatz function. Additional participant encouragement is also highly advised - we suggest more direct forms of compensation such as a monetary payment. Rewarding participants for better guesses should also be considered to improve the data quality. However, with more rewards for better results, cheating should be taken into account. 

For dataset creation, it is advised to use fully private datasets of over 200 sentences, created specifically for this purpose. This eliminates cheating by lookup, and leaves only possible cheating by using a language model, a method that is likely too complicated for the average participant. If this is infeasible, news sampling should consist of at least 20 articles over wider time periods, resulting in 200-300 sentences. 

For LLM usage, the core concerns to be addressed are modeling representative training and test sets. This raises the question of what constitutes a language, and what parts of a language are more important. This is left for future work.

Finally, the published experiment results enable some additional analysis that is out of scope for this work, such as examining how guessing accuracy varies by position within a word or by character. 

\section*{Ethical Considerations}
\subsection*{Use of Volunteers}
Participants for the experiment were recruited using word-of-mouth and social media channels. Each participant was informed about the nature of the experiment prior to providing consent and no personal data was collected other than an optional request to be named in acknowledgments. 

\subsection*{Use of AI-based Writing Assistance}
During this study, large language models of the Claude family were used extensively for research, code generation, and proofreading purposes. All of the text was written by the authors. 

\section*{Acknowledgements}

We thank the following volunteers for contributing their time and effort to advancing Ukrainian linguistics:
007morf, Alina Kryvobokova, Anastasiia Loban, Andrii, AnimeGame, Artem Dzhemesiuk, Artem Sokolik, BeHappy1337, Bob, Bohdan Rubakha, BrokenByteOfCode, DARIA, Danylo Vorvul, DitriX, Dmytro Potapov, Kateryna Hordiienko, KroZen\_Dev, Maksbid, Mamontovozik, Mariia Chepeliuk, Sofiia Kuplevatska, Viktor Kauk, Vladis002, Yan Krasnyi, grantuseyes, lime, marmazon, moonlightqrl, popo, rawrshah, virusebe, Yevhen Domeretskyi, Yevhen Movsesov, Yevhen Tuturov, Yevhenii, Yevhenii, Yevhenii, Yevhenii Batiuta, Yehor, Yelyzaveta, Ivan, Ivan, Ivan, Ivan Turchyn, Ihor, Ihor, Ihor Braichenko, Illia Repin, Inna Kaliafitska, Iryna, Aleks, Aleks, Alinochka, Anastasiia, Anastasiia Pysmak, Anatolii, Anatolii, Anhelion, Andrii, Andrii B, Andrii Voinarovskyi, Andrii L., Andrii Liashchuk, Anna, Anton, Artem, Artem, Artem, Artem Olimpiiev, Artur, Artur Kornilov, Bairaktar, Vadym, Valeriia, Vlad Milka, Vladyslav, Vladyslav, Vladyslav Diachenko, Vladyslav Zamotailo, Volodymyr, Volodymyr Yatsynych, Vik Lisovyi, Viktoriia, Viktoriia, Vitalii, Vitalii, Havrylov Illia, Herman, Herman, Hladka Marharyta, Davyd4, Dmytro, Dmytro, Dmytro, Dmytro S, Dmytro Ch, Zhanna, Zlata, Kartoplianchyk, Kateryna, Kateryna Tatarchuk, Kolesnyk Vlad, Kostiantyn, Kisilov Kyrylo, Kit, Larysa, Liubov, Maksym, Maksym, Marharyta, Markiian, Mariia, Mykyta, MykytaYe, Mykola, Mykola, Mykola Usik, Mykolai, Mykhailo, Mykhailo, Mohyla Stanislav, Miroshnykov Illia, Nazarii, Nataliia, Nataliia Hrechushkina, Nataliia Puniak, NeuroKit, Nikita, Oleh, Oleh, Oleksandr, Oleksandr, Oleksandr Bohun, Oleksandr D, Oleksandr and Iryna, Oleksandra, Oleksii Maryshchyn, Olena, Olha, Olha, Pavlo, Pavlo Chuiko, Roman P., Roman Serdiuk, Rina, Rinat Marinad, Riia Kovalenko, Svizhyi, Serhii, Serhii O., Sonia, Stesi, Tamila Krashtan, Taras, Yuliia, Yurii, Yurii, Yurii Prokopenko, Yaroslav, Yaroslav, and an additional 25 people who have decided to stay anonymous.

We thank Yurii Paniv and Mariana Romanyshyn for valuable discussion that helped shape this work, and NeuroKit for invaluable assistance with volunteer recruitment.

\bibliography{references}

\appendix
\section{Published Data}

We publish the code and data used in this work, as well as the raw data gathered from volunteers. The results of the experiment are anonymized.

Experiment and processing code: \href{https://github.com/arimlabs/ule}{https://github.com/arimlabs/ule}

Dataset: \href{https://huggingface.co/datasets/a-l-o/shortnews}{https://huggingface.co/datasets/a-l-o/shortnews}

Data gathered: \href{https://huggingface.co/datasets/a-l-o/ule_results}{https://huggingface.co/datasets/a-l-o/ule\_results}

\end{document}